\documentclass{article}

\usepackage{arxiv}

\usepackage[utf8]{inputenc} 
\usepackage[T1]{fontenc}    
\usepackage{hyperref}       
\usepackage{url}            
\usepackage{booktabs}       
\usepackage{amsfonts}       
\usepackage{nicefrac}       
\usepackage{microtype}      
\usepackage{lipsum}
\usepackage{multirow}
\usepackage{makecell}
\usepackage{amsmath}
\usepackage{amssymb}
\usepackage{multirow}
\usepackage{subfigure,graphicx}
\usepackage{courier}
\usepackage{booktabs}
\usepackage{latexsym,amsmath,amssymb,graphicx}
\usepackage{url}
\usepackage{color}
\usepackage{bm}
\usepackage{helvet}
\usepackage{algorithm}
\usepackage{algorithmic}
\newcommand{\etal}{\textit{et al}. }
\newcommand{\ie}{\textit{i}.\textit{e}., }

\title{Fusformer: A Transformer-based Fusion Approach for Hyperspectral Image Super-resolution}

\author{
	Jin-Fan Hu\\
	School of Mathematical Sciences\\
	University of Electronic Science and Technology of China\\
	Chengdu, 611731, China\\
	\texttt{hujf0206@163.com}
		\And
	Ting-Zhu Huang\\
	School of Mathematical Sciences\\
	University of Electronic Science and Technology of China\\
	Chengdu, 611731, China\\
	\texttt{tingzhuhuang@126.com}   
	\And
  Liang-Jian Deng\\
  School of Mathematical Sciences\\
  University of Electronic Science and Technology of China\\
  Chengdu, 611731, China\\
  \texttt{liangjian.deng@uestc.edu.cn}   
}

\begin{document}
\maketitle

\begin{abstract}
Hyperspectral image has become increasingly crucial due to its abundant spectral information. However, It has poor spatial resolution with the limitation of the current imaging mechanism. Nowadays, many convolutional neural networks have been proposed for the hyperspectral image super-resolution problem. However, convolutional neural network (CNN) based methods only consider the local information instead of the global one with the limited kernel size of receptive field in the convolution operation. In this paper, we design a network based on the transformer for fusing the low-resolution hyperspectral images and high-resolution multispectral images to obtain the high-resolution hyperspectral images. Thanks to the representing ability of the transformer, our approach is able to explore the intrinsic relationships of features globally. Furthermore, considering the LR-HSIs hold the main spectral structure, the network focuses on the spatial detail estimation releasing from the burden of reconstructing the whole data. It reduces the mapping space of the proposed network, which enhances the final performance. Various experiments and quality indexes show our approach's superiority compared with other state-of-the-art methods.
\end{abstract}


\section{Introduction}

Hyperspectral images have recently received a lot of attention in many fields since the rich spectral information of each pixel in a scenario provides more abundant and reliable knowledge than those multispectral images (e.g., RGB images). As a result, many researchers start taking advantage of the characteristics of HSI in various computer vision tasks, such as classification, segmentation, and object tracking begin the HSIs for better performance. However, limited by the current physical imaging system, there is an unavoidable trade-off, the high spatial resolution and high spectral resolution cannot be obtained at the same time \cite{dian2020recent}. The imaging mechanism can only capture the image with high spatial resolution along with limited spectral bands, e.g., high-resolution multispectral image (HR-MSI), or low spatial resolution but with a higher spectral resolution, e.g., low-resolution hyperspectral image (LR-HSI) in practice. Thus, Fusing the high spatial resolution of MSI and high spectral resolution of HSI becomes a promising technique to generate the desired high-resolution hyperspectral image (HR-HSI).

Many methods have been proposed from various perspectives to address the HSI super-resolution problem in the last few years. They are roughly divided into two classes: traditional methods and deep learning-based methods. As for the traditional methods, many researchers introduce different prior knowledge in their proposed optimization models for exploiting the intrinsic properties under the maximum a posteriori (MAP) framework. 

Since matrix or tensor factorization-based methods show reliable and promising performance in many computer vision and image processing tasks \cite{jiang2020framelet,8948303,7579662,Jiang_2017_CVPR}, many researchers also introduce the matrix or tensor factorization methodology into the hyperspectral image super-resolution or pansharpening problems \cite{han2017hyperspectral,GLP-HS,1518950,CSTF,CNN-FUS,UTV,pan2019multispectral,LTMR,LTTR,Wu,9106801,deng2018variational,deng2019fusion}.
Although those methods have achieved excellent performance, they often require known or estimated $ \mathbf{B} $ and $ \mathbf{R} $ beforehand, which are difficult to obtain in practice. Furthermore, the representation ability of those handcraft regularization terms is usually limited in actual life, and their optimal parameters need to be tuned for different devices.

Recently, deep learning-based methods have shown satisfactory performance in different fields due to their remarkable feature representing ability. Thus, many researchers take the deep learning, especially the CNN technique into consideration for fusing the LR-HSI and HR-MSI \cite{MHFnetpami,dian2018deep,shao2018remote,li2017hyperspectral,vitale2019cnn,han2019hyperspectral,liu2018deep,huang2015new,liu2018psgan,CNN-FUS,SSRNET,ResTFNet} which have outperformed many traditional methods. Nevertheless, due to its insufficient information extraction ability of convolution, there is still room for improvement.

In this work, we notice that the transformer and its various modifications have obtained outstanding achievements in natural language processing and computer vision tasks \cite{vaswani2017attention,chen2021pre,dosovitskiy2020image}. Hence, a transformer-based network architecture called Fusformer is proposed for the hyperspectral image super-resolution problem. Our method integrates a self-attention mechanism that can exploit more global relationships among the data than the convolution operation with a limited receptive field. 
Furthermore, we force our Fusformer to estimate the residuals instead of reconstructing the whole HR-HSI, enabling the network architecture to learn in a smaller mapping space.
This paper designs an efficient network architecture to solve the HSI super-resolution problem via fusing the HR-MSI and LR-HSI. To sum up, the contributions of this paper are presented as follows:
\begin{enumerate}
	\item To our knowledge, it is the first time using the transformer to solve the hyperspectral image super-resolution problem. The self-attention mechanism in the transformer enables our network to represent more global information than previous CNN architectures.
	
	\item The proposed approach focuses on the residual domain instead of the primitive image domain, which leads to a smaller mapping space. It relieves the network of the burden of reconstructing the desired HR-HSI directly.
	
	\item Only a few parameters are involved in the network with light computation making our approach practical in the real-life application. Furthermore, the network is plain and easy to follow. Thus, future researches can be improved based on our simple yet effective architecture.
\end{enumerate}

The organization of this article is as follows. Section II will introduce our Fusformer architecture in detail. In Section III, extensive experiments on different datasets are presented to validate the superiority of our network. Finally, we draw the conclusion in Section IV. In this paper, we utilize non-bold case, bold upper case, and calligraphic upper case letters to denote the scalar, matrix, tensor, respectively.

\section{Network Architecture}
\label{sec:NA}
With the rapid development of deep learning techniques, CNN-based approaches are also used for solving many tasks of computer vision and image processing, including the pansharpening and the HSI-MSI fusion problem\cite{MHFnetpami,dian2018deep,shao2018remote,li2017hyperspectral,vitale2019cnn,han2019hyperspectral,liu2018deep,huang2015new,liu2018psgan, CNN-FUS, SSRNET, ResTFNet}. They have obtained state-of-the-art performance in recent years due to their powerful feature extraction ability.
Notwithstanding the remarkable achievements of those CNN-based methods have obtained, the core elements in the neural network are those various convolution kernels with localized kernel sizes. Thus, the region of interest by convolution is restricted within a small area, \ie the convolution operation is conducted locally, and the global structure is neglected while it contains valuable information. Considering the limitation of the convolution, how to better extract and understand the global information becomes a difficult but vital issue. 
\subsection{Background}
The transformer model was created by Vaswani \etal in 2017\cite{vaswani2017attention} to collect better long-term information than recurrent neural networks and convolutional neural networks. The proposed transformer outperforms other methods and has been proven to be quite crucial in natural language processing (NLP) tasks. Furthermore, motivated by the success of transformer architecture in NLP, Dosovitskiy \etal\cite{dosovitskiy2020image} propose the vision transformer (ViT) for image classification and Chen \etal\cite{chen2021pre} design the image processing transformer (IPT) to address low-level vision tasks. Both of them obtain the best results compared to existing approaches. The achievements of the transformer in various computer vision tasks inspire us to design a network based on the transformer to solve the hyperspectral image super-resolution problem via the superior ability to capture long-term information and relationships.
\begin{figure*}[hbtp]
	\begin{center}
		\begin{minipage}{ 1\linewidth}
			\centering
			\includegraphics[width=0.9\linewidth]{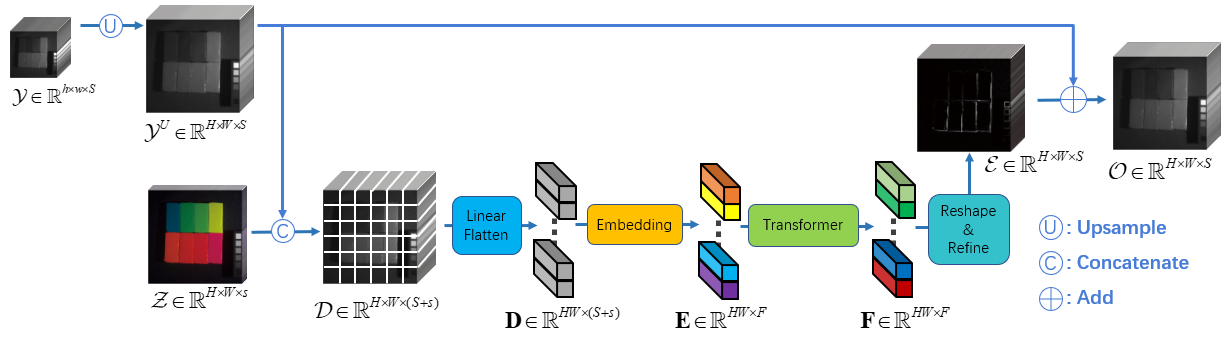}
			\centering
			
			{(a)}
		\end{minipage}
		
		\begin{minipage}{0.7\linewidth}
			\centering
			\includegraphics[width=1\linewidth]{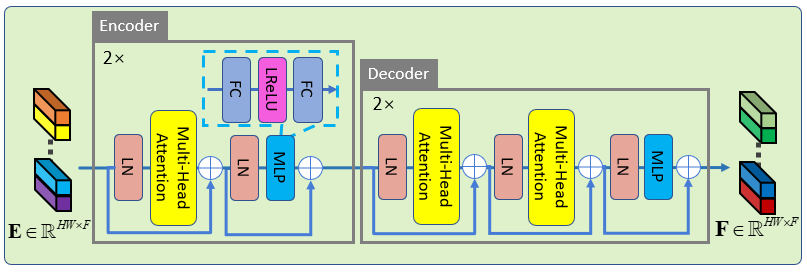}
			\centering
			
			{(b) Transformer} 
		\end{minipage}
		\hspace{7pt}
		\begin{minipage}{0.2\linewidth}
			\begin{minipage}{1\linewidth}
				\centering
				\includegraphics[width=1\linewidth]{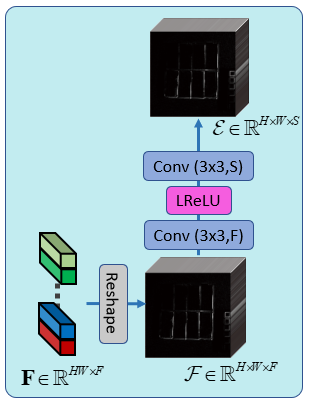}
				\centering
				
				{(c) Reshape \& Refine}
			\end{minipage}
		\end{minipage}
		
		\caption{Overview of the proposed Fusformer network. (b) The transformer model used in the network. (c) An illustration of the reshape \& refine module.}\label{Flowchart}
	\end{center}
\end{figure*}

%
%
\subsection{Proposed Method}
The global information is barely used due to the limitation of the regular convolution operation with the local region of interest in the CNN architecture. Hence, we expect to use the transformer model to consider the features and information effectively and globally. The flowchart of our network structure is presented in Fig. \ref{Flowchart}. Note that the inputs of our Fusformer are the upsampled LR-HSI $\mathcal{Y}^{U} \in \mathbb{R}^{H \times W \times S}$ and the HR-MSI $\mathcal{Z} \in \mathbb{R}^{H \times W \times s}$ since the $\mathcal{Y}^{U}$ holds the similar structure as the ground-truth HR-HSI $\mathcal{X}$. 
\subsubsection{Input of Transformer}
The HR-MSI $\mathcal{Z}$ is then concatenated with $\mathcal{Y}^{U}$ along the spectral dimension. After obtaining the data cube $\mathcal{D} \in \mathbb{R}^{H \times W \times (S+s)}$ containing the raw spectral and spatial information, we unfold the tensor $\mathcal{D}$ to a matrix $\mathbf{D} \in \mathbb{R}^{HW \times (S+s)}$, due to the input's dimension requirement of the transformer model. It is worth noting that each vector in the matrix $\mathbf{D}$ has its physical meaning \ie representing a pixel in the image with the spectral structure and corresponding spatial information. While other transformer-based methods for computer vision tasks\cite{dosovitskiy2020image, chen2021pre} reshape a small image patch into a vector directly. On the one hand, the hyperspectral image contains more spectral bands than other natural RGB images. The vector reshaped from an image patch will be too long to compute. On the other hand, we believe a vector representing pixel information instead of a patch is also suitable for our pixel-wise super-resolution problem. Hence, the transformer model is quite consistent with the characteristics of the hyperspectral image super-resolution issue. Every pixel can be naturally represented as a vector, and the transformer architecture enables the network to discover and consider the relationships among all the pixels globally. With a simple fully connected layer, the matrix $\mathbf{D} \in \mathbb{R}^{HW\times (S+s)}$ is then embedding to the matrix $\mathbf{E} \in \mathbb{R}^{HW\times F}$, where $F$ denotes the number of feature channels ($F = 48 $ in this paper.). 
Next, we send the embedded patches to the transformer model, which is represented in Fig. \ref{Flowchart}-(b).
\subsubsection{Transformer}
The transformer model is the main part of our architecture which is shown in Fig. \ref{Flowchart}-(b). Here we use both the encoder and decoder part of the original transformer. As for the encoder (See the top of Fig. \ref{Flowchart}-(b)), LN indicates the layer normalization which is widely used in the transformer-based methods \cite{vaswani2017attention,chen2021pre,dosovitskiy2020image} for the training's stability. Multi-head attention is the self-attention mechanism with multi-heads that enables the network to capture and consider the relationship globally. 
The general procedures of the self-attention mechanism are as follows. 
\begin{equation}\label{eq:sa}
\begin{aligned}
\mathbf{Q}  = &\mathbf{W}_q\mathbf{X},\mathbf{K}  = \mathbf{W}_k\mathbf{X}, \mathbf{V}  = \mathbf{W}_v\mathbf{X},\\
Attention &= \operatorname{softmax}(\frac{\mathbf{XW}_q\mathbf{W}_k^T\mathbf{X}^T}{\sqrt{d_k}})\mathbf{XW}_v \\
&= \operatorname{softmax}(\frac{\mathbf{QK}^T}{\sqrt{d_k}})\mathbf{V},
\end{aligned}
\end{equation}
where $\mathbf{W}_q$, $\mathbf{W}_k$ and $\mathbf{W}_v$ denote the corresponding weight matrices need to be trained, $d_k$ represents the dimension of $\mathbf{K}$ for scaling. As for the multi-head attention, $L$
heads denotes $L$ individual groups of ($\mathbf{Q}_i, \mathbf{K}_i, \mathbf{V}_i$) ($i = 1,2,\cdots,L$) with $L$ attention values. Furthermore,  $\operatorname{softmax}$ gives attention values from 0 to 1 which differentiate levels of importance in \textbf{V}.
The whole algorithm of the encoder can be described as follows.
\begin{equation}\label{eq:en}
\begin{aligned}
\mathbf{X}'  &= \operatorname{MHA}(\operatorname{LN}(\mathbf{X}))\\
\mathbf{X} &= \mathbf{X}+\mathbf{X}' \\
\mathbf{X}'  &= \operatorname{MLP}(\operatorname{LN}(\mathbf{X}))\\
\mathbf{X} &= \mathbf{X}+\mathbf{X}',\\
\end{aligned}
\end{equation}
where $\operatorname{LN}$ represents the layer normalization,  $\operatorname{MHA}$ denotes the multi-head attention module and the $\operatorname{MLP}$ defines the multi-layer perception plotted in Fig. \ref{Flowchart}-(c).
Similarly, the decoder can be described as follows.
\begin{equation}\label{eq:de}
\begin{aligned}
\mathbf{X}'  &= \operatorname{MHA}(\operatorname{LN}(\mathbf{X}))\\
\mathbf{X} &= \mathbf{X}+\mathbf{X}' \\
\mathbf{X}'  &= \operatorname{MHA}(\operatorname{LN}(\mathbf{X}))\\
\mathbf{X} &= \mathbf{X}+\mathbf{X}' \\
\mathbf{X}'  &= \operatorname{MLP}(\operatorname{LN}(\mathbf{X}))\\
\mathbf{X} &= \mathbf{X}+\mathbf{X}' \\
\end{aligned}
\end{equation}
After getting the learned features $\mathbf{F} \in \mathbb{R}^{HW \times F}$, we reshape it back to a 3D tensor $\mathcal{F} \in \mathbb{R}^{H\times W \times F}$ and then feed it to a refine module for the desired residual $\mathcal{E} \in \mathbb{R}^{H \times W \times S}$. Finally, the output $\mathcal{O}$ is obtained by adding the upsampled LR-HSI $\mathcal{Y}^U$ and learned residual $\mathcal{E}$.

\begin{figure*}[ptb]
	\centering
		\begin{center}
	{\includegraphics[width=1\linewidth]{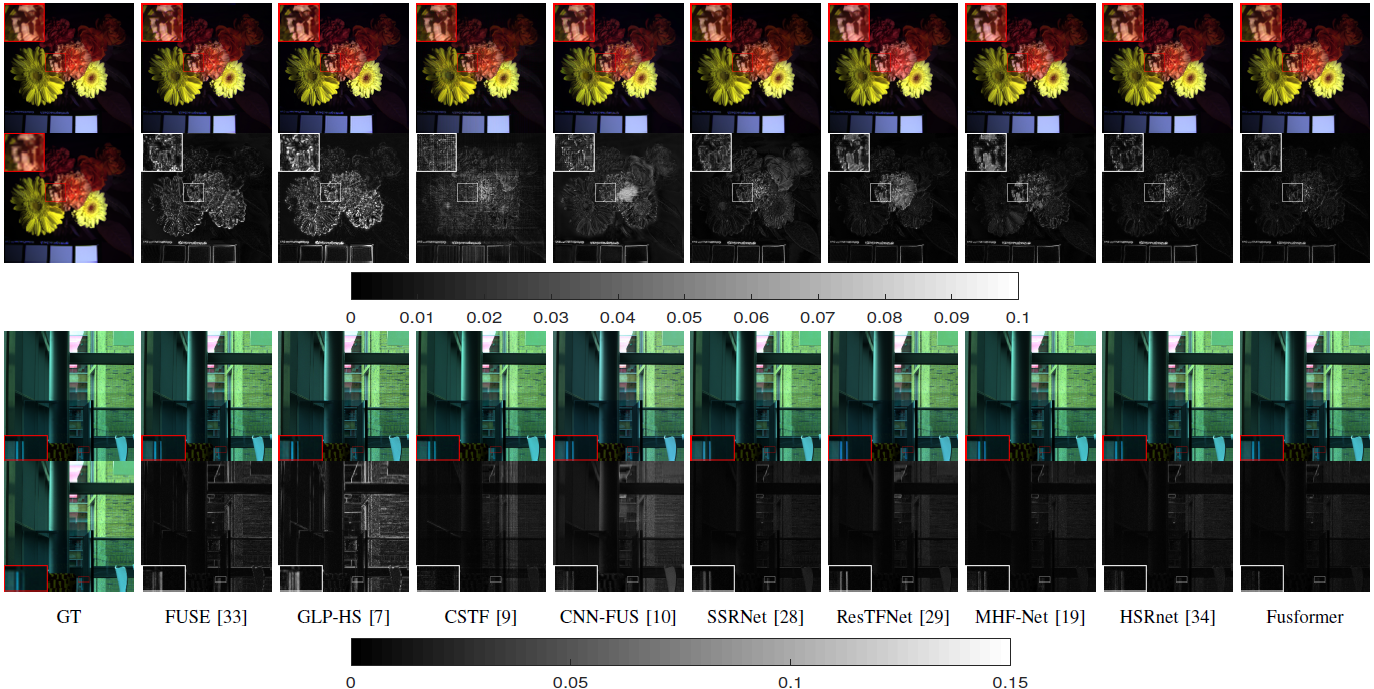}}
		\end{center}
	
	\caption{The first column: the true pseudo-color images from the CAVE and Harvard dataset and the corresponding LR-HSI images of \textit{feathers} (R-31, G-24, B-2) (1st-2nd rows) and \textit{window} (R-29, G-21, B-13) (3th-4th rows). The 2nd-8th columns: the true pseudo-color fused products and the corresponding residuals for the different methods in the benchmark pointing out some close-ups to facilitate the visual analysis.}
	\label{F:cave}
\end{figure*}

\section{Experiment Results}
To verify the effectiveness of our proposed Fusformer, we compare with classic and state-of-the-art 1) traditional approaches: FUSE \cite{FUSE}, GLP-HS \cite{GLP-HS}, CSTF \cite{CSTF} and CNN-FUS \cite{CNN-FUS}. 2) deep learning-based approaches: SSRNet \cite{SSRNET}, ResTFNet \cite{ResTFNet}, MHF-Net \cite{MHFnetpami} and HSRnet \cite{Hutnnls} on two different hyperspectral image datasets \ie CAVE dataset \cite{yasuma2010generalized} and Harvard dataset \cite{chakrabarti2011statistics}.
Both of them contain hyperspectral images with 31 spectral channels. Images of CAVE dataset are with a size of 512 $\times$ 512, while images of Harvard dataset are cropped with the spatial size of 1000 $\times$ 1000. 20 images from CAVE dataset are selected for training, 11 images from CAVE dataset, and 10 images from Harvard dataset are for testing. Note that all the training images are from the CAVE dataset. The images from the Harvard dataset thus can be used for the generalization test, which is quite crucial for deep learning-based methods. 
\begin{table}[h]
	\centering\renewcommand\arraystretch{1}\setlength{\tabcolsep}{5pt}
	\caption{Average QIs of the results on 11 testing images from the CAVE dataset and the corresponding number of parameters. The best values are highlighted in boldface, the second best values are highlighted in underline, while M indicates a million. }		
	\begin{tabular}{lccccc}
		\Xhline{1.2pt}
		Method& PSNR & SAM & ERGAS & SSIM &\# parameters\\ \hline
		FUSE\cite{FUSE} & 39.72 & 5.83 & 4.18 & 0.975 & $/$\\ 
		GLP-HS\cite{GLP-HS} & 37.81 & 5.36 & 4.66 & 0.972 & $/$\\ 
		CSTF\cite{CSTF} & 42.14 & 9.92 & 3.08 & 0.964 & $/$\\ 
		CNN-FUS\cite{CNN-FUS} & 42.66 & 6.44 & 2.95 & 0.982& $/$ \\ 
		SSRNet\cite{SSRNET} & 45.28 & 4.72 & 2.06 & 0.990& \textbf{0.03M} \\
		ResTFNet\cite{ResTFNet} & 45.35  & 3.76 & 1.98 & \underline{0.993}& 2.26M \\
		MHF-Net\cite{MHFnetpami} & 46.32 & 4.33 & 1.74 & 0.992& 3.63M \\
		HSRnet \cite{Hutnnls}  & \underline{47.82} & \underline{2.66} & \underline{1.34} & \textbf{0.995} &1.90M \\ 
		Fusformer & \textbf{48.56} & \textbf{2.52} & \textbf{1.30} & \textbf{0.995}& \underline{0.10M} \\ \hline
		Best value& +$ \infty $ & 0 & 0 & 1 & 0 \\ 
		\Xhline{1.2pt}
	\end{tabular}
	\label{cave11-ave}
\end{table}

Tab. \ref{cave11-ave} and \ref{harvard10-ave} list the quantitative comparisons in CAVE and Harvard datasets. Our proposed Fusformer obtains the best results on almost every QI and involves only 0.1 million parameters, making our network more practical. We also show visual presentation and corresponding absolute residual maps of two samples selected from CAVE and Harvard dataset in Fig. \ref{F:cave}. It is obvious that our approach still outperforms other methods and the residuals are the darkest.
\subsection{Generalization Ability}
Furthermore, deep learning-based methods usually perform poorly on examples that differ from the training dataset. Hence, the generalization ability of deep learning-based methods is quite crucial.
However, our Fusformer is still satisfying, and only the ERGAS is not the smallest. The generalization performance of HSRnet is close to the proposed Fusformer, but its involved parameters are much more than Fusofrmer.

\begin{table}[h]
	\centering\renewcommand\arraystretch{1}\setlength{\tabcolsep}{5pt}
	\caption{Average QIs of the results for 10 testing images from the Harvard dataset and the corresponding number of parameters. The best values are highlighted in boldface, the second best values are highlighted in underline, while M indicates a million.}
	\begin{tabular}{lccccc}
		\Xhline{1.2pt}
		Method 	& PSNR & SAM & ERGAS & SSIM &\# parameters\\ \hline
		
		FUSE\cite{FUSE} & 42.06 & \underline{3.23} & 3.14 & 0.977 &$/$\\ 
		GLP-HS\cite{GLP-HS} & 40.14 & 3.52 & 3.74 & 0.966 &$/$\\
		CSTF\cite{CSTF} & 42.97 & 3.30 & \textbf{2.43} & 0.972 &$/$\\  
		CNN-FUS\cite{CNN-FUS} & 43.61 & 3.32 & 2.78 & \underline{0.978} &$/$\\ 
		SSRNet\cite{SSRNET} & 39.87 & 5.40 & 5.44 & 0.963 &\textbf{0.03M}\\
		ResTFNet\cite{ResTFNet} & 38.39 & 5.85 & 6.98 & 0.957 &2.26M\\ 
		MHF-Net\cite{MHFnetpami} &  40.37 & 4.64 & 24.17 & 0.966 & 3.63M\\
		HSRnet\cite{Hutnnls} &  \underline{44.28} & \textbf{2.66} & \underline{2.45} & \textbf{0.984}& 1.90M\\
		Fusformer & \textbf{44.42} & \textbf{2.66} & 2.48 & \textbf{0.984} &\underline{0.10M}\\ 
		\hline
		Best value& +$ \infty $ & 0 & 0 & 1 &0\\ \Xhline{1.2pt}
	\end{tabular}
	\label{harvard10-ave}
\end{table}
\subsection{Ablation Study}
Our Fusformer is excepted for learning the residuals between the upsampled LR-HSI $\mathcal{Y}^U$ and ground-truth $\mathcal{X}$ instead of reconstructing the whole HSI. We conduct a simple experiment to verify the residual learning strategy (RLS) \ie adding the upsampled LR-HSI $\mathcal{Y}^U$ to the outcome learned by the network. Tab. \ref{T:hp} records the results of the architecture with or without the RLS. It is clear that adding the upsampled LR-HSI $\mathcal{Y}^U$ is quite vital for the network. The rough information of $\mathcal{Y}^U$ helps the network to boost the performance and strengthen the stability.

\begin{table}[h]
	\centering\footnotesize
	\renewcommand\arraystretch{0.9}\setlength{\tabcolsep}{6pt}
	\caption{Average QIs and related standard deviations of the results on the CAVE dataset using the proposed method with and without the residual learning strategy. The best values are highlighted in boldface. }\label{T:hp}
	\begin{tabular}{lccccc}
		\Xhline{1.2pt}
		Method 		& PSNR 	& SAM 	& ERGAS & SSIM \\ \hline
		W/o RLS	& 42.71$\pm$7.82& 3.29$\pm$1.24 & 4.48$\pm$7.43 & 0.984$\pm$0.02 \\
		Fusformer & \textbf{48.56}$\pm$3.03& \textbf{2.52}$\pm$0.83 & \textbf{1.30}$\pm$0.86& \textbf{0.995}$\pm$0.00\\ \Xhline{1.2pt}
	\end{tabular}	
\end{table}

\section{Conclusion}
In this work, a transformer-based network architecture called Fusformer is proposed. Compared with previous CNN-based methods, our method can consider the global information instead of the local information in a receptive field with a limited kernel size. This is the first time adopting the transformer model in the hyperspectral image super-resolution issue to the best of our knowledge. Our method is simple yet effective and contains few parameters. Future researches can further exploit the potential base on our proposed framework.


\section{Acknowledge}
The work is supported by National Natural Science Foundation of China grants 12001446, 61702083, 12171072 and 61876203, and the Fundamental Research Funds for the Central Universities JBK2001011.

\bibliographystyle{unsrt}  


\bibliography{references1}

\end{document}